\documentclass[sigconf]{acmart}
\AtBeginDocument{%
  }

\usepackage{graphicx}
\usepackage{float}

\begin{document}

\title{LinkedIn Post Embeddings: Industrial Scale Embedding Generation and Usage across LinkedIn}

\author{Sudarshan Srinivasa Ramanujam}
\authornote{All authors contributed equally to this research.}
\affiliation{%
  \institution{LinkedIn Corporation}
  \city{Mountain View}
  \state{CA}
  \country{USA}}
\email{sramanujam@linkedin.com}

\author{Akanksha Bindal}
\authornotemark[1]
\affiliation{%
  \institution{LinkedIn Corporation}
  \city{Mountain View}
  \state{CA}
  \country{USA}}
\email{abindal@linkedin.com}

\author{Yu Jiang}
\authornotemark[1]
\affiliation{%
  \institution{LinkedIn Corporation}
  \city{Mountain View}
  \state{CA}
  \country{USA}}
\email{tjiang@linkedin.com}

\author{Timothy J. Hazen}
\authornotemark[1]
\affiliation{%
  \institution{LinkedIn Corporation}
  \city{Mountain View}
  \state{CA}
  \country{USA}}
\email{thazen@linkedin.com}

\author{David Golland}
\authornotemark[1]
\affiliation{%
  \institution{LinkedIn Corporation}
  \city{Mountain View}
  \state{CA}
  \country{USA}}
\email{dgolland@linkedin.com}

\author{Fengyu Zhang}
\affiliation{%
  \institution{LinkedIn Corporation}
  \city{Mountain View}
  \state{CA}
  \country{USA}}
\email{fezhang@linkedin.com}

\author{Daqi Sun}
\affiliation{%
  \institution{LinkedIn Corporation}
  \city{Mountain View}
  \state{CA}
  \country{USA}}
\email{daqsun@linkedin.com}

\author{Wanning Li}
\affiliation{%
  \institution{LinkedIn Corporation}
  \city{Mountain View}
  \state{CA}
  \country{USA}}
\email{wannli@linkedin.com}

\author{Birjodh Singh Tiwana}
\affiliation{%
  \institution{LinkedIn Corporation}
  \city{Mountain View}
  \state{CA}
  \country{USA}}
\email{btiwana@linkedin.com}

\author{Siddharth Dangi}
\affiliation{%
  \institution{LinkedIn Corporation}
  \city{Mountain View}
  \state{CA}
  \country{USA}}
\email{sdangi@linkedin.com}

\author{Peng Yan}
\authornote{LinkedIn Alumni}
\affiliation{%
  \institution{LinkedIn Corporation}
  \city{Mountain View}
  \state{CA}
  \country{USA}}
\email{pyan@linkedin.com}

\renewcommand{\shortauthors}{Sudarshan Srinivasa Ramanujam et al.}
\settopmatter{printacmref=false} 
\renewcommand\footnotetextcopyrightpermission[1]{} 
\pagestyle{plain} 
\begin{abstract}
A post embedding (representation of text in embedding space that effectively captures semantic meaning) is a foundational component of LinkedIn that is consumed by product surfaces in retrieval and ranking (e.g., ranking posts in the feed or video tab). This paper presents the post embeddings used at LinkedIn, where a pre-trained transformer-based large language model (LLM) is taken as input and fine-tuned using multi-task learning across a diverse set of semantic labeling tasks.  We observe positive transfer, leading to improved performance across all tasks, compared to training them independently. The generated post embeddings outperform baseline models in zero-shot learning, demonstrating its potential for broader applicability. Furthermore, the generated post embeddings' performance surpasses that of OpenAI’s ADA-001 and ADA-002 embeddings on LinkedIn specific datasets and tasks. We also describe the offline evaluation methodology and the deployment to our near-line infrastructure, which makes the post embedding available for use within minutes of post creation for any downstream application. We present how the embeddings were applied in the Feed product surface, in both ranking and retrieval stages, and showcase the real world online impact to demonstrate the superior performance of these embeddings. Finally, we also share the results of applying the embeddings to the retrieval system of our video ranking product surface in LinkedIn. These embeddings have been battle-tested in production at LinkedIn for over two years, consistently powering multiple products.
\end{abstract}

\maketitle
\begin{center}
\small\itshape
Accepted for presentation at the 34th ACM International Conference on Information and Knowledge Management (CIKM 2025).\\
\end{center}

\section{Introduction}
LinkedIn is the world’s largest professional network, connecting over a billion users across 200+ countries and territories 
\cite{borisyuk2024lignn,borisyuk2024lirank}. Our platform fosters a thriving information exchange ecosystem, helping members discover valuable content, learn new skills, and explore career opportunities. Among various content types, posts play a crucial role in facilitating knowledge sharing between creators and consumers. To enhance content discovery and engagement, LinkedIn introduced out-of-network content recommendations, significantly expanding the pool of candidate posts. This shift made efficient embedding-based retrieval (EBR) across a vast corpus essential for delivering high-quality recommendations. Consequently, state-of-the-art recommendation models require rich, semantically meaningful representations of posts to improve content understanding and ranking. Prior work has explored various approaches for converting text into embeddings that effectively capture semantic meaning. These methods range from traditional word vector models like Word2Vec and GloVe to more advanced transformer-based models such as BERT\cite{devlin2018bert}, RoBERTa \cite{liu2019roberta}, T5  \cite{zoupanos2022efficient} and more recently, to OpenAI’s models such as ADA-002 \cite{greene2022new}. These newer OpenAI embedding models are designed for out-of-the-box usage, requiring no fine-tuning, and are intended to perform effectively in zero-shot learning scenarios.
In this paper, we introduce LinkedIn Post Embedding model which is LinkedIn’s content understanding model, designed to generate high-quality post embeddings that power multiple downstream applications, including Feed ranking, Feed retrieval, out-of-network recommendations and immersive video experiences at a low dimensionality of 50. We also share how these embeddings are leveraged in a few downstream applications in a production setting for retrieval and ranking and demonstrate the utility of having embeddings that can capture semantics.

\subsection{Key Contributions}

\begin{itemize}
    \item \textbf{Model Architecture for LinkedIn Post Embedding} – A multi-task fine-tuning approach for training LLMs on diverse datasets to generate post embeddings.
    \item \textbf{Offline Evaluation Methodology} - A semantic understanding metric to measure embedding quality.
    \item \textbf{Online Deployment} - Design for making embeddings accessible across LinkedIn’s product surfaces.
    \item \textbf{Offline Results} - Offline results after multi-task fine-tuning including positive transfer among tasks and comparison against OpenAI embeddings on LinkedIn benchmarks. 
    \item \textbf{Online Impact in Feed Ranking and Retrieval} - Evaluation of the real world effectiveness of LinkedIn post embeddings after deployment on the LinkedIn feed platform. We share practical examples of how embeddings can be integrated into ranking and retrieval models and the impact online through A/B testing. In this work, we share feed ranking, feed retrieval and video retrieval as examples.  
\end{itemize}

\section{Related Work}
Pre-trained models such as BERT, RoBERTa, and more
recently, models like GPT-3 \cite{brown2020language} and E5 \cite{wang2022text}, have demonstrated remarkable performance improvements across a variety of Natural
Language Processing (NLP) tasks. Recent
studies have shown the effectiveness of fine-tuning
models on multiple tasks to achieve better generalization
and performance across all tasks. Aghajanyan et al.
(2021) introduced MUPPET \cite{aghajanyan2021muppet}, which demonstrated that
pre-fine-tuning on a diverse set of tasks could
significantly improve the model's performance on
individual tasks. Contrastive learning has emerged as
a powerful technique for training embeddings by
distinguishing between similar and dissimilar pairs.
Reimers and Gurevych (2019) proposed Sentence-BERT,
which trains both a siamese network architecture and a triplet architecture to generate
embeddings for sentences, significantly improving
performance on sentence similarity tasks \cite{reimers2019sentence}. Despite
these advances, learning approaches can be
sensitive to the quality of positive and negative pairs, and
obtaining high-quality labeled data can be challenging.
Models like XLM-R have shown
that fine-tuning on multilingual data can lead to robust
cross-lingual embeddings \cite{conneau2019unsupervised}. Greene et al. (2022)
explored the performance of OpenAI's ADA embeddings,
highlighting its performance on a multitude of tasks, underscoring its generalized nature \cite{greene2022new}.
However, these models often face limitations when
applied directly to specific domains such as
recommendation systems due to the linguistic variability
between the general pre-trained embeddings and the
specialized application domain. This requires additional fine-tuning to achieve optimal performance in domain-specific tasks. In our investigation, we encountered several challenges when trying to
incorporate widely available architectures into production
environments. These challenges included smaller context
windows, limited linguistic variability in the topic
ontology, and high embedding dimension size resulting in
increased latency for production scale recommendation
systems.
In this paper, we build upon these foundational works by
implementing a multi-task learning approach
tailored to LinkedIn's unique content. By leveraging
diverse semantic labeling tasks, we enhance the
model's semantic understanding, improve multilingual support, and achieve competitive performance with significant compression. Our goal is to provide valuable insights into developing and deploying specialized embeddings in large-scale, real-world applications.

\section{Vision for Training Platform}
\begin{itemize}
    \item \textbf{Accuracy}: the embeddings should accurately capture all of the relevant information about a post so that they can reliably be leveraged by downstream applications to make predictions about engagement or decisions about distribution. 
    \item \textbf{Robustness}: the embedding should be able to handle all inputs relevant to the post, including text in any language used on LinkedIn. The training data did not have any filters based on language.
    \item \textbf{Timeliness}: to ensure the embeddings reflect the latest trends and patterns in the data, we should have the ability to retrain as often as needed
    \item \textbf{Extensibility}: Adding new sources of data or new tasks to the training pipeline needs to be standardized for any teams in LinkedIn to adopt or contribute data sources for fine-tuning.
    \item \textbf{Flexibility}: It should be easy to experiment with different underlying architectures for various base language models since this space is rapidly evolving. 
    \item \textbf{Easy Deployability}: Seamless integration with required offline/nearline systems to ramp to production as quickly as possible. 
\end{itemize}

\section{Datasets}
\label{sec:dataset}

\begin{table}[H]
\centering
\small
\begin{tabular}{|l|p{0.72\columnwidth}|}
\hline
\textbf{Dataset} & \textbf{Description} \\
\hline
\textbf{Interest} & Derived from LinkedIn’s topic tagging models, which classify posts into categories based on a structured ontology. A pair of posts is labeled positive if they share the same interest category. \\
\hline
\textbf{Storyline} & Editor-curated posts grouped by topics. Available in 50+ languages, this adds multilingual posts to the training data. These can be typically seen in the top right section of the LinkedIn feed in desktop\\
\hline
\textbf{Hashtag} & Uses post hashtags as soft labels. This is available in all languages on LinkedIn platform.  \\
\hline
\textbf{Search} & Extracts query-post relevance pairs from LinkedIn’s content search data. \\
\hline
\textbf{Intent} & Classifies posts based on intent (e.g., share advice, job seeking, motivation). \textbf{Used for evaluation to assess zero-shot generalization and not used for training the model.} \\
\hline
\end{tabular}
\caption{Training and evaluation datasets overview}
\label{tab:datasets}
\end{table}

\section{Modeling Architecture}
In this section we first describe the single task training setup used for training content embeddings followed by the multi-task set up.

In our work, we have multiple tasks and multiple datasets. However, each data set is used for only one task. We did try other losses (margin maximization loss, prediction tasks) but none of them performed as well as the siamese architecture which we will illustrate in the subsections below. 

\subsection{Single Task Architecture}

\begin{figure}
    \centering
    \includegraphics[width=0.9\linewidth]{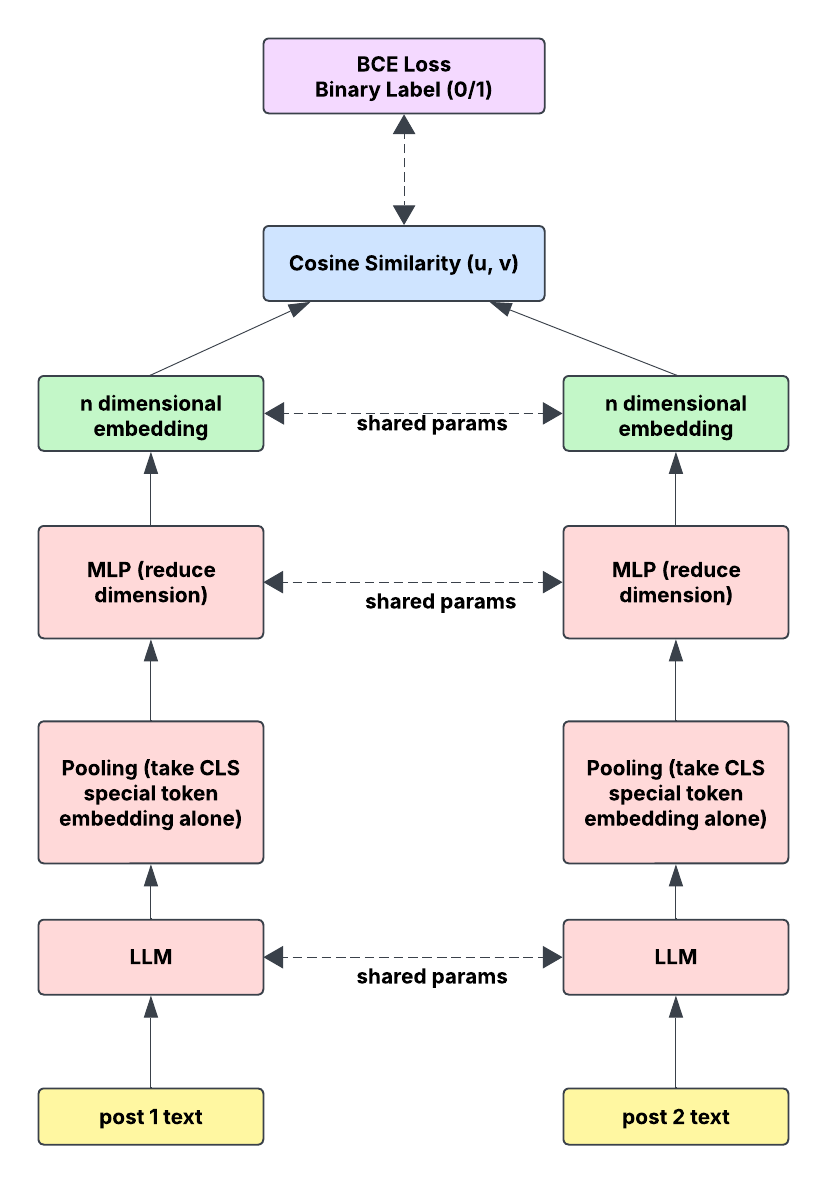}
    \caption{Architecture used for a single task}
    \Description{Architecture used for a single task}
    \label{fig:single_task_arch}
\end{figure}

\autoref{fig:single_task_arch} represents the siamese architecture which is a representation of a single task that was replicated across datasets for training a multi-task model \cite{reimers2019sentence}. We collect pairs of posts from the dataset (both positives and negatives). Positive examples are sampled in different ways for each dataset and the negative examples are two randomly sampled posts for every dataset (more details in \autoref{sec:dataset}).

\begin{itemize}
    \item Positive pairs (label = 1): P1 and P2 are topically
related and should produce embeddings that have a high cosine similarity.
Example: P1 and P2 are about bitcoin.
\item Negative pairs (label = 0): P1 and P2 are NOT
related and should produce embeddings that have a low cosine similarity.
Example: P1 is about ML, P2 is about sports.
\end{itemize} 

The label assigned is binary (1 or 0) and we apply a binary cross entropy loss between the label and the cosine similarity of the embeddings of the two posts.

\begin{equation} \label{cosine_bce_loss}
\begin{aligned}
\mathcal{L} = - \frac{1}{N} \sum_{i=1}^{N} \big[ 
& y_i \cdot \log\left( \sigma\left( \cos(\mathbf{e}_{1}^{(i)}, \mathbf{e}_{2}^{(i)}) \right) \right) \\
& + (1 - y_i) \cdot \log\left( 1 - \sigma\left( \cos(\mathbf{e}_{1}^{(i)}, \mathbf{e}_{2}^{(i)}) \right) \right) 
\big]
\end{aligned}
\end{equation}

Where:
\begin{itemize}
    \item $N$ is the number of pairs.
    \item $y_i \in \{0, 1\}$ is the binary label for the $i$-th pair.
    \item $\mathbf{e}_{1}^{(i)}$, $\mathbf{e}_{2}^{(i)}$ are the embeddings for the $i$-th pair of posts.
    \item $\cos(\mathbf{e}_1, \mathbf{e}_2) = \frac{\mathbf{e}_1 \cdot \mathbf{e}_2}{\|\mathbf{e}_1\| \|\mathbf{e}_2\|}$ is the cosine similarity between two embeddings.
    \item $\sigma(x) = \frac{1}{1 + e^{-x}}$ is the sigmoid function.
\end{itemize}
\subsection{Multi-Task Architecture}

\begin{figure}
    \centering
    \includegraphics[width=0.9\linewidth]{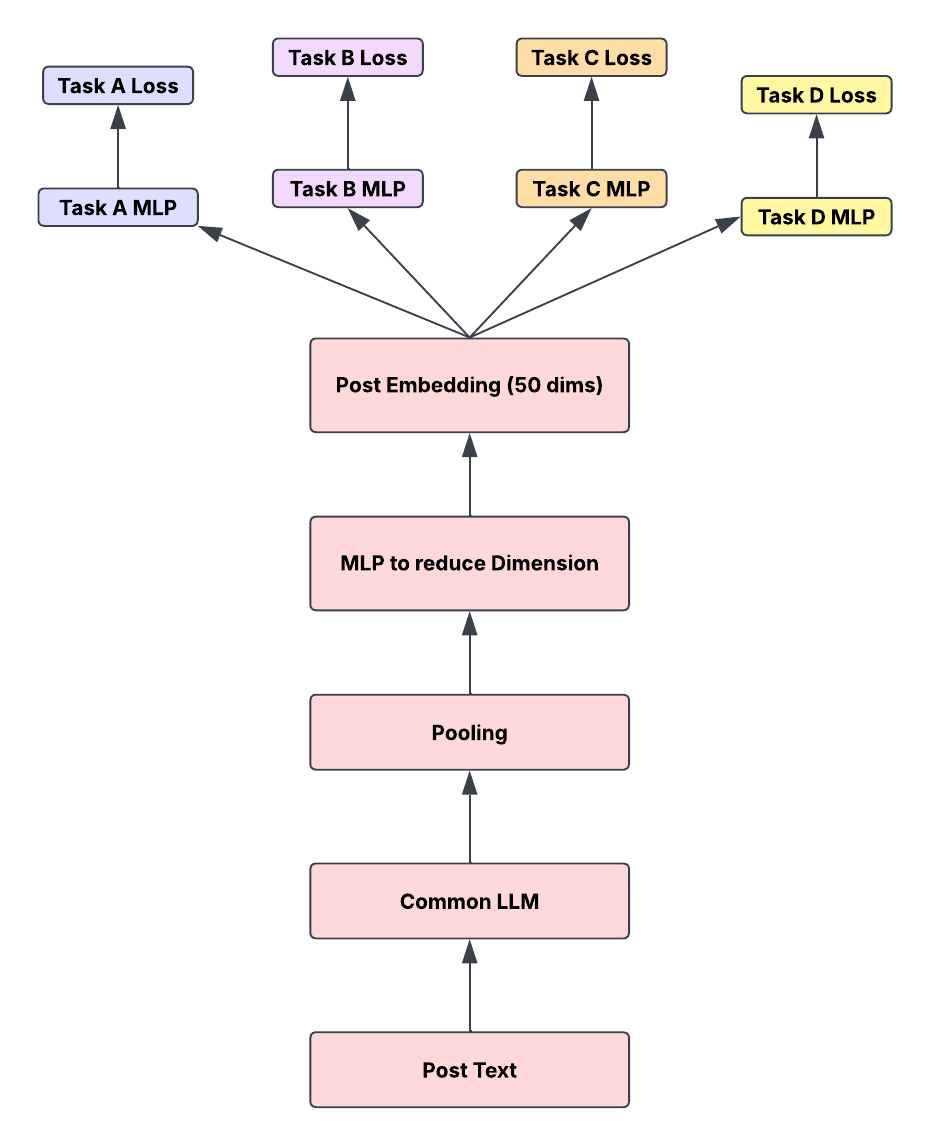}
    \caption{Multitask architecture}
    \Description{Multitask training set up architecture}
    \label{fig:multi_task_arch}
\end{figure}

We expand the single task setup to a multi-task training paradigm. The key idea of using a multi-task approach is that adding new datasets/tasks helps all the tasks being trained \cite{aghajanyan2021muppet} versus training one model for every isolated task. The single task architecture described in the previous section is duplicated for multiple datasets (each dataset has its own independent task). We simultaneously train for several tasks with shared LLM parameters, allowing effective semantic representation to be efficiently learned within a single model. Each independent task tower computes its own loss. In our implementation \textbf{each task was an independent siamese task with BCE loss} similar to the set up in the single task architecture.  
With this approach we can fine-tune an LLM that has awareness of the semantics required for consumption by multiple downstream product teams. The parameters for the common LLM and the layer to reduce dimension are shared across all tasks. On completion of training, the task level heads are removed and the shared layers are used to infer the post embeddings. \autoref{fig:multi_task_arch} illustrates how the multi-task training is set up.  The output of the [CLS] special token is employed as the pooling methodology to get the output from the LLM prior to dimension reduction. 

\subsubsection{Task Heterogeneous Sampling}

\begin{figure}
    \centering
    \includegraphics[width=\linewidth]{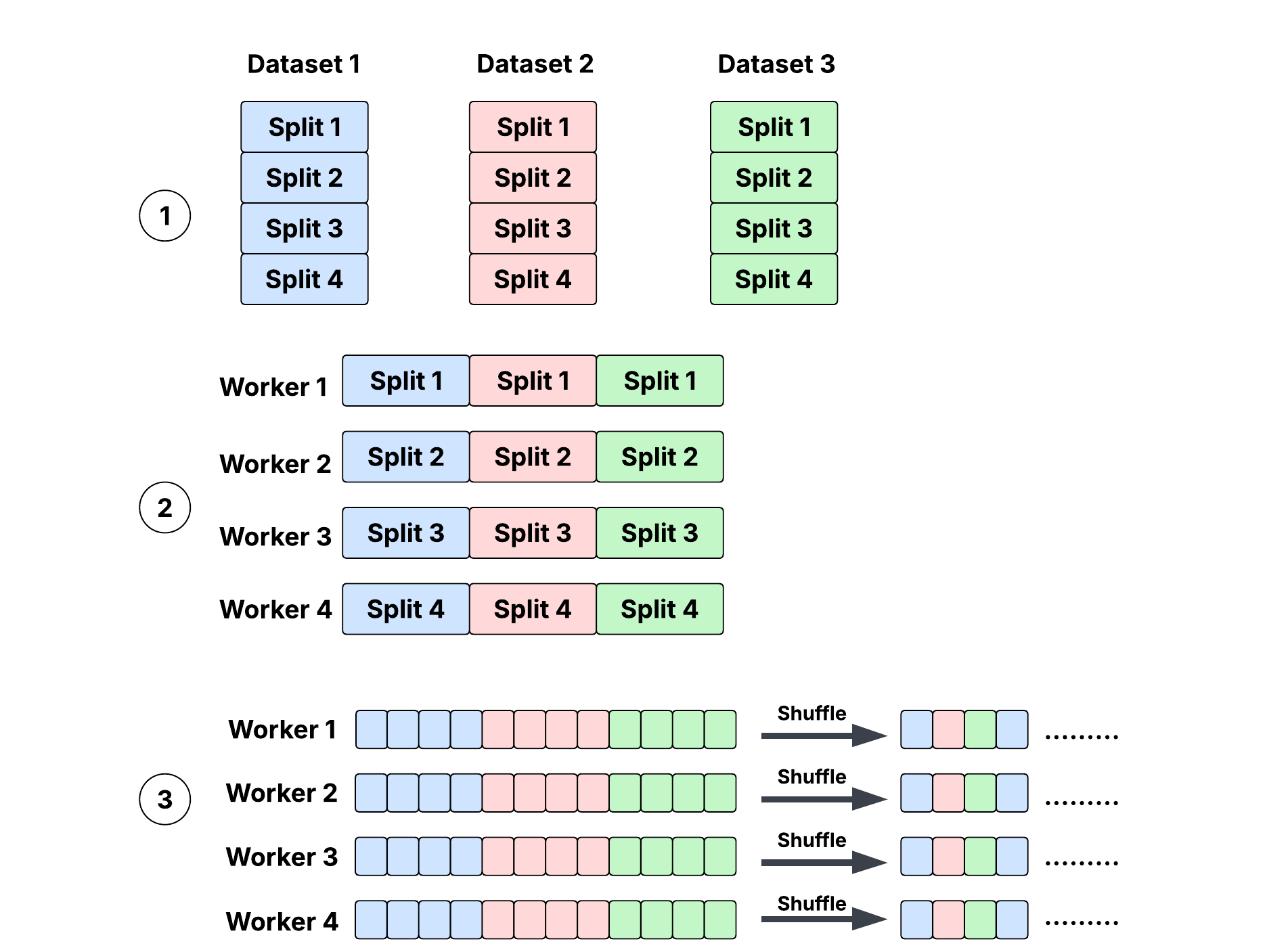}
    \caption{Task heterogeneous sampling with 3 datasets and 4 workers}
    \Description{Task heterogeneous sampling}
    \label{fig:task_heterogeneous_sampling}
\end{figure}

Similar to the implementation in the muppet paper \cite{aghajanyan2021muppet}, we sample data from all task data sources within one batch which helps with training stability and in gradient steps being better balanced across the tasks.  \autoref{fig:task_heterogeneous_sampling} illustrates the three steps below.

\begin{enumerate}
    \item For each dataset, randomly split it into the number of workers’ splits.
    \item For each worker, load its corresponding split for all datasets.
    \item First batch data on each split, then randomly shuffle all the batches.
\end{enumerate}

With this approach every batch consists of data from multiple datasets.  The final loss employed was the average loss across all tasks. In the event of significant skew in dataset volume, a weighting term for each task could be added to help with the training. For this work, we did not add any weighting to the loss.

\subsection{Implementation Details}

We used 104M training samples coming from a combination
of datasets described in \autoref{sec:dataset}. We use a 6 layer multilingual BERT (pre-trained on LinkedIn data using masked-language modeling) as the base model \cite{devlin2018bert}, with a total
parameter size of 89M and vocabulary size of 135K. We
use 1 worker and 6 GPUs for training. We use a per GPU batch size of 32 for siamese fine-tuning and shared
embedding size of 50. We selected an embedding dimension of 50 after empirical experimentation. 
This dimension provides a balance between expressiveness and latency in large scale deployment. 
Higher dimensions achieved only marginally better offline accuracy while incurring higher inference cost and storage requirements. For task level parameters, each task has an MLP layer of size (50x100).
We use a learning rate of 1e-6 for training. All experiments were conducted on a CentOS Linux server equipped with dual Intel\textsuperscript{\textregistered} Xeon\textsuperscript{\textregistered} Silver~4216 
(Cascade Lake) CPUs (32~cores, 2.10~GHz), 64~GB of RAM, and an NVIDIA Tesla V100 
SXM2 GPU with 32~GB memory, using CUDA Toolkit~11.7.

\subsection{Member Embeddings from Post Embeddings}

Numerous applications in LinkedIn require an embedding representation of LinkedIn members, and extracting this representation in an efficient way is essential.
For example, any EBR (embedding-based retrieval) application needs both query (member) and item embeddings that are in the same space. 

We made use of hierarchical clustering (Ward's method)\cite{pal2020pinnersage} to get the representations of the member (medoids) based on member engagement in Feed and post embeddings of those engagements. \begin{enumerate}
    \item Identify engagement history (e.g., like, comment, share, react)
    \item Join post embeddings to the corresponding history in engagement
    \item Run hierarchical clustering (Ward's method) to generate topK medoids for every member. 
\end{enumerate}
The top 'K' is based on an importance score for each cluster which is a combination of size of the cluster and freshness of items in the cluster \cite{pal2020pinnersage}.

\section{Offline Evaluation}

Training produces a model that captures post
semantics that we evaluate across different
downstream tasks. For instance, 2 posts on deep learning should be close in the embedding space. We built a simulation of EBR offline to evaluate the embeddings. We build a dataset of triplets containing anchor, positive and negative texts. The expectation is that for each anchor, the positive item is as close as possible and the negative item is as far as possible in the embedding space.

\begin{table}[h]
    \centering
    \small
    \begin{tabular}{c c c}
        \hline
        \textbf{Anchor} & \textbf{Positive} & \textbf{Negatives} \\
        \hline
        $a_1$ & $p_1$ & $\{n_{11}, n_{12}, \dots, n_{1N}\}$ \\
        $a_2$ & $p_2$ & $\{n_{21}, n_{22}, \dots, n_{2N}\}$ \\
        $\vdots$ & $\vdots$ & $\vdots$ \\
        $a_M$ & $p_M$ & $\{n_{M1}, n_{M2}, \dots, n_{MN}\}$ \\
        \hline
    \end{tabular}
    \caption{Anchor-Positive-Negative triplets}
    \label{tab:triplets}
\end{table}

After training a candidate embedding model, we generate embeddings for all the text in the evaluation dataset, and then calculate the average fraction of triplets, where the distance between the anchor and positive instance is smaller than the anchor and negative instance. This serves as a good proxy for embedding based retrieval applications and is used as the offline evaluation metric for our content models.

\textbf{AvgFracTripletsWherePosIsCloser}:
Fraction of triplets where the positive is closer to the anchor than the negative (larger is better):

\begin{equation} \label{triplet_eval_equation}
\frac{1}{MN} \sum_{i=1}^{M} \sum_{j=1}^{N} 
\begin{cases} 
1, & \text{if } \text{dist}(a_i, p_i) < \text{dist}(a_i, n_{ij}) \\ 
0, & \text{otherwise} 
\end{cases}
\end{equation}

\section{Online Deployment}
\begin{figure}
    \centering
    \includegraphics[width=\linewidth]{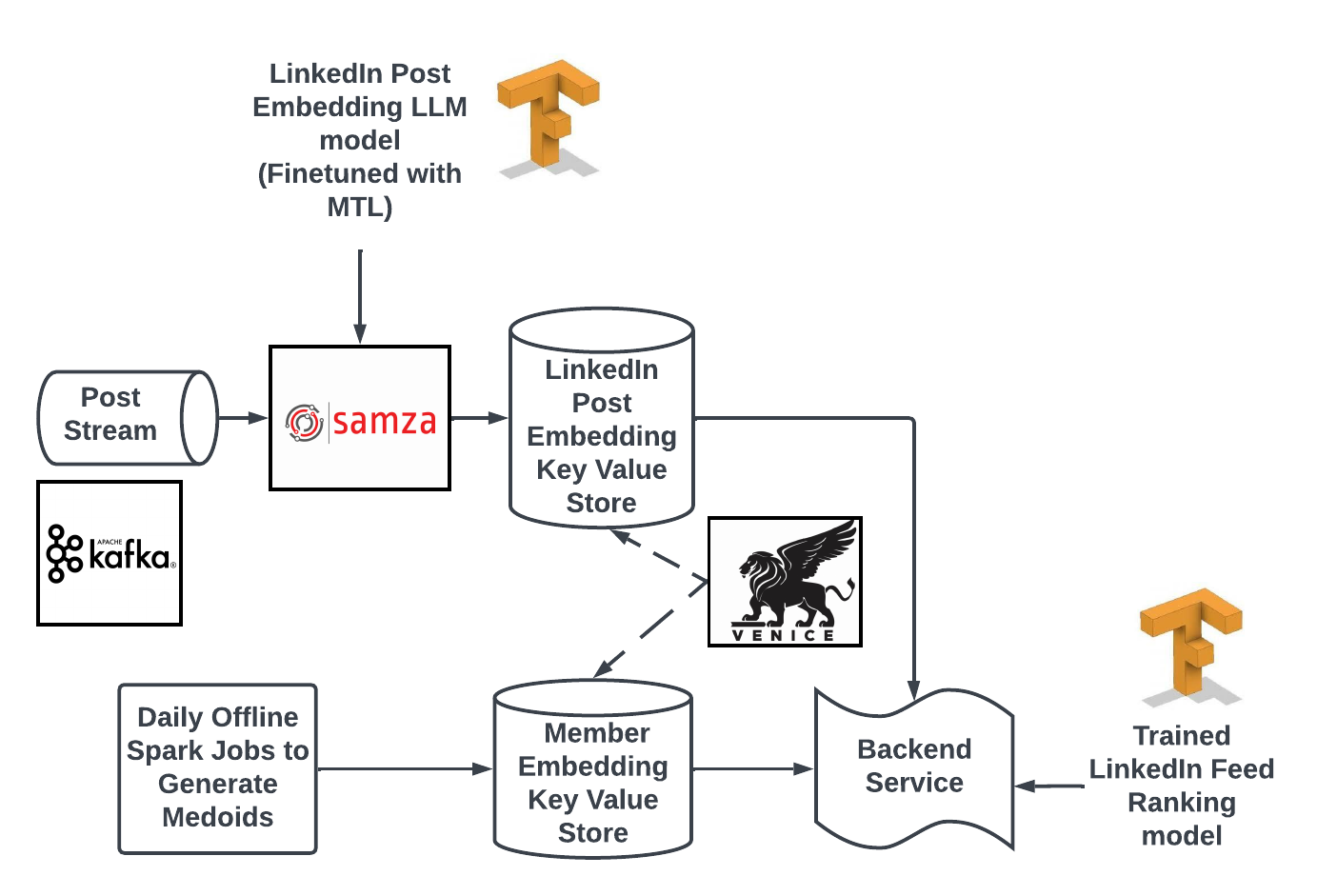}
    \caption{Online system for post embeddings}
    \Description{Online system for post embeddings}
    \label{fig:lipost_online_arch}
\end{figure}

\autoref{fig:lipost_online_arch} shows the high level overview of the online system we use in the Feed ranking model. All incoming posts that are created, are fed into a Samza job that computes embeddings using the trained LinkedIn post embedding model, and pushes it to a key-value store within 2 mins of post creation (typically this is done in a matter of seconds). The posts' key-value store is configured to store a fixed history of embeddings at all times on a rolling basis. Any backend service can fetch the embedding feature for scoring. Derived member embeddings (medoids), are pushed to a dedicated key-value store and this is an offline job which runs once a day. 

\section{Results}

In \autoref{result_subsec_1} to \autoref{result_subsec_3} we will talk about offline results. To ensure fair evaluation, we report results on complete test data sets rather than sampled subsets. Multiple retraining runs produced only marginal differences in offline evaluation metrics, so we report representative results for clarity. For all of our online A/B tests, we report the statistical significance of our results in \autoref{subsec:res_impact_downstream}. The performance improvements of our multi-task model are consistent across multiple datasets and tasks, reinforcing the robustness of our approach. 

\subsection{Fine-tuning an LLM on multiple tasks at once helps uplift performance in all tasks} 
\label{result_subsec_1}

\begin{table}[h]
    \centering
    \small
    \setlength{\tabcolsep}{4pt} 
    \begin{tabular}{l c c c}
        \hline
        \textbf{Model} & \textbf{E1} & \textbf{E2} & \textbf{E3} \\
        & (Interest) & (Story) & (Hashtag) \\
        \hline
        T1 (Interests) & 0.88 & 0.86 & 0.79 \\
        T2 (Story) & 0.76 & 0.93 & 0.85 \\
        T3 (Hashtag) & 0.79 & 0.93 & 0.93 \\
        \textbf{LinkedIn post embedding (MTL)} & \textbf{0.89} & \textbf{0.95} & \textbf{0.93} \\
        \hline
    \end{tabular}
    \caption{Evaluation results across models using AvgFracTripletsWherePosIsCloser; E1-E3 correspond to different evaluation datasets}
    \label{tab:evaluation_all_tasks_help}
\end{table}

T1, T2, and T3 correspond to models trained only on the independent datasets, and the last row is the LinkedIn Post Embedding model, trained in a multi-task fashion with all the datasets. E1, E2 and E3 are eval datasets built using the corresponding dataset mentioned in \autoref{tab:evaluation_all_tasks_help} (Interest, Storyline and Hashtag). 
The results demonstrate that our model trained on a combination of data from multiple semantic labeling tasks, shows a better overall performance across all tasks. The first 3 rows serve as an ablation study showing the impact with the use of equivalent dataset only as opposed to MTL framework. Content search data was used for training but was not used for evaluation purposes, since there were no immediate plans to deploy these embeddings to the content search surface.

\subsection{Zero Shot Capabilities Improvements}

\begin{table}[h]
    \centering
    \small
    \setlength{\tabcolsep}{4pt} 
    \begin{tabular}{l c}
        \hline
        \textbf{Model} & \textbf{E4 (Intent)} \\
        \hline
        T4 (Intent) & 0.69 \\
        \textbf{LinkedIn Post Embedding (MTL)} & \textbf{0.72} \\
        \hline
    \end{tabular}
    \caption{Evaluation results for intent understanding (E4) using AvgFracTripletsWherePosIsCloser}
    \label{tab:intent_evaluation}
\end{table}
\autoref{tab:intent_evaluation} demonstrates zero shot learning capabilities for the post embedding model. Although it is trained only on data from Interests, Search, Storylines and Hashtag datasets, it generalizes effectively to the Post Intent Dataset (E4). On Task E4, LinkedIn Post Embeddings outperforms the model fine-tuned solely on T4 (Post Intent training data).




\subsection{Comparing performance with generalized OpenAI embeddings}
\label{result_subsec_3}

\textbf{E1} - Interest Dataset
\textbf{E2} - Storyline Dataset
\textbf{E3} - Hashtag Dataset

\begin{table}[h]
    \centering
    \small
    \setlength{\tabcolsep}{4pt} 
    \begin{tabular}{l c c c c}
        \hline
        \textbf{Model} & \textbf{Dim} & \textbf{E1} & \textbf{E2} & \textbf{E3} \\
        \hline
        BERT-base & 768 & 0.69 & 0.90 & 0.77 \\
        ADA\_001 & 1024 & 0.66 & 0.95 & 0.82 \\
        ADA\_002 & 1536 & 0.89 & 0.95 & 0.89 \\
        E5-base-v2 & 768 & 0.84 & 0.96 & 0.87 \\
        E5-multilingual-base & 1024 & 0.81 & 0.96 & 0.87 \\
        \textbf{LinkedIn Post Embedding} & \textbf{50} & \textbf{0.89} & \textbf{0.95} & \textbf{0.93} \\
        \hline
    \end{tabular}
    \caption{Performance comparison across models (including ADA\_002 \cite{greene2022new}) using AvgFracTripletsWherePosIsCloser}
    \label{tab:model_performance}
\end{table}

The results in Table 5 show that compared to open-source models that generate generalized embeddings, we achieve comparable performance with up to 30x compression in embedding size for LinkedIn specific tasks.


\subsection{Impact in Downstream Applications}
\label{subsec:res_impact_downstream}

All test results reported here are based on online A/B experiments run for at least one week, and all downstream application impacts are statistically significant with \textbf{p value less than 0.05}.
\subsubsection{Feed Ranking:}

\begin{figure}
    \centering
    \includegraphics[width=\linewidth]{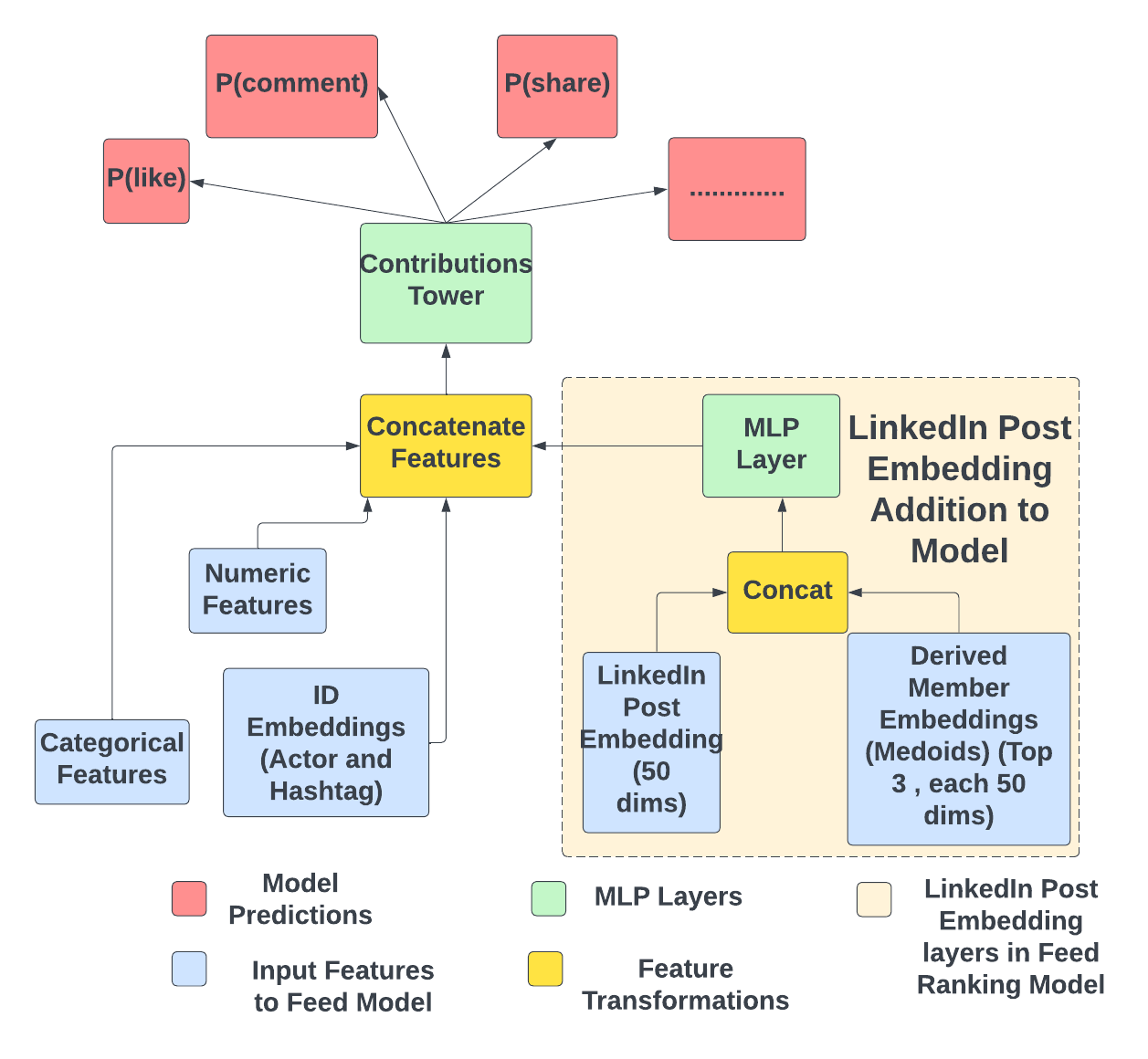}
    \caption{Feed ranking model architecture with LinkedIn post embeddings}
    \Description{Feed ranking model architecture with LinkedIn post embeddings}
    \label{fig:feed_spr_model_arch}
\end{figure}

The Feed ranking model is the final ranking layer, which takes in inputs from multiple first pass rankers (examples: followed content,  jobs content, suggested content, etc.), and outputs the final ranked list. The current Feed ranking model is a large personalized model which includes ID features, numeric features and categorical features \cite{borisyuk2024lirank}. 
\autoref{fig:feed_spr_model_arch} shows how the embeddings were integrated into the main Feed ranking model. 

After adding LinkedIn Post Embeddings to the model, we were able to achieve an increase of \textbf{0.1\% in the number of user sessions} on LinkedIn ($p < 0.001$). This metric indicates that more members found the feed more relevant and chose to come back more often. We achieved an increase of \textbf{0.21\% in the number of daily unique professional interactions} by our members ($p < 0.0001$) and \textbf{revenue} was up by \textbf{0.42\%} ($p < 0.001$) in online A/B tests. While these lifts may appear modest, at LinkedIn's scale, this translates to millions of additional positive member interactions daily, representing a significant business impact.

\subsubsection{Feed Retrieval:}

\begin{figure}
    \centering
    \includegraphics[width=\linewidth]{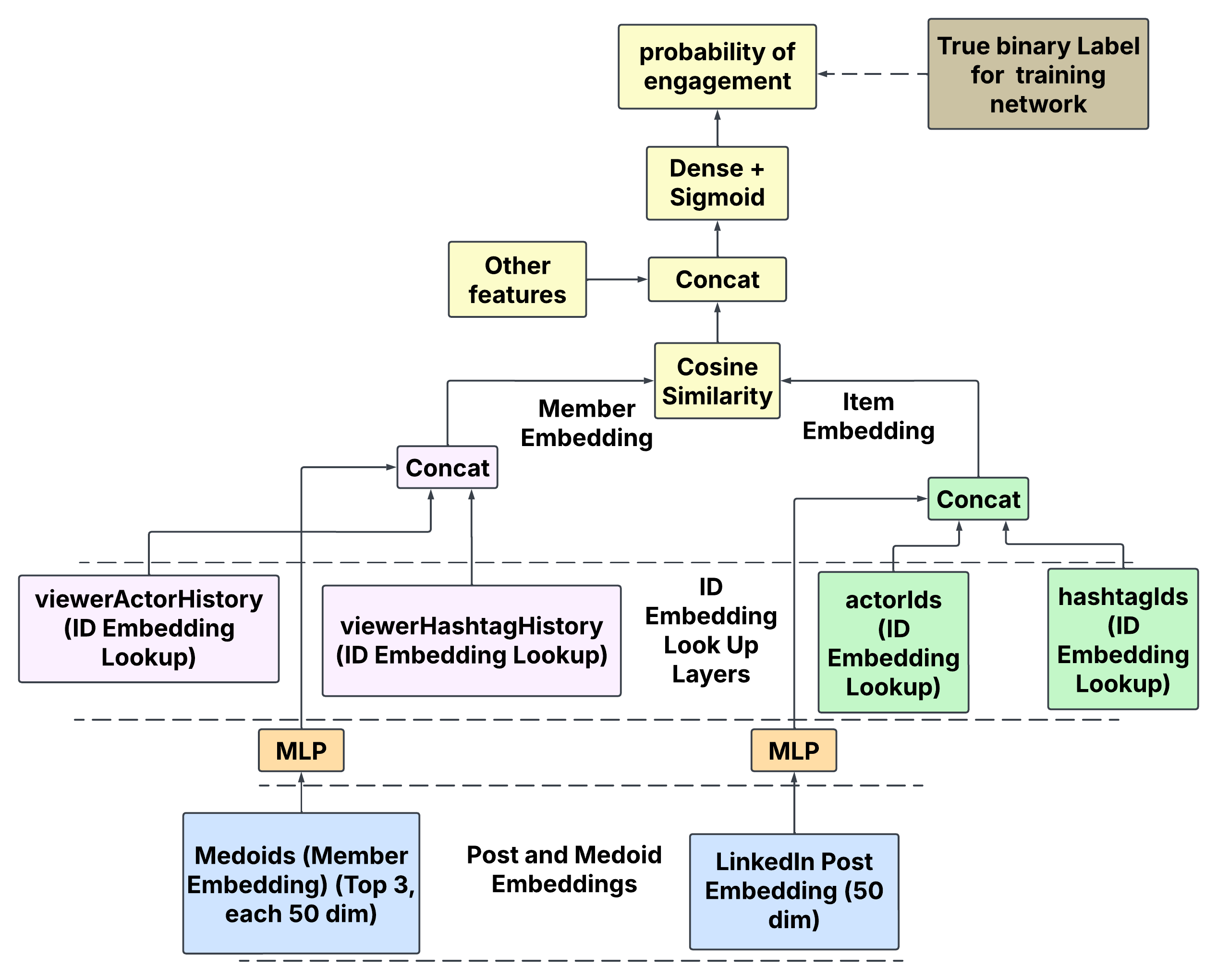}
    \caption{Feed retrieval model architecture with LinkedIn post embeddings}
    \Description{Feed retrieval model architecture with LinkedIn post embeddings}
    \label{fig:feed_retrieval_model_arch}
\end{figure}

The Feed model has a retrieval layer for fetching the best candidates for ranking from the corpus of content created by a member's connections. This is one of the sources that feeds into the final ranking model along with out-of-network posts, videos, ads, jobs and others. In this layer, we filter down top 500 most relevant connected content for the user from a corpus which could range up to millions depending on connection size of a member. See \autoref{fig:feed_retrieval_model_arch} for the architecture of the retrieval model after addition of LinkedIn Post Embeddings. The model resulted in a \textbf{+0.37\% increase in daily unique members who had an active engagement in the Feed} ($p < 0.001$), and a \textbf{0.05\% decrease in Feed skips, indicating greater relevance} ($p < 0.0001$).

\subsubsection{Feed Video Recommendation:}

\begin{figure}
    \centering
    \includegraphics[width=0.95\linewidth]{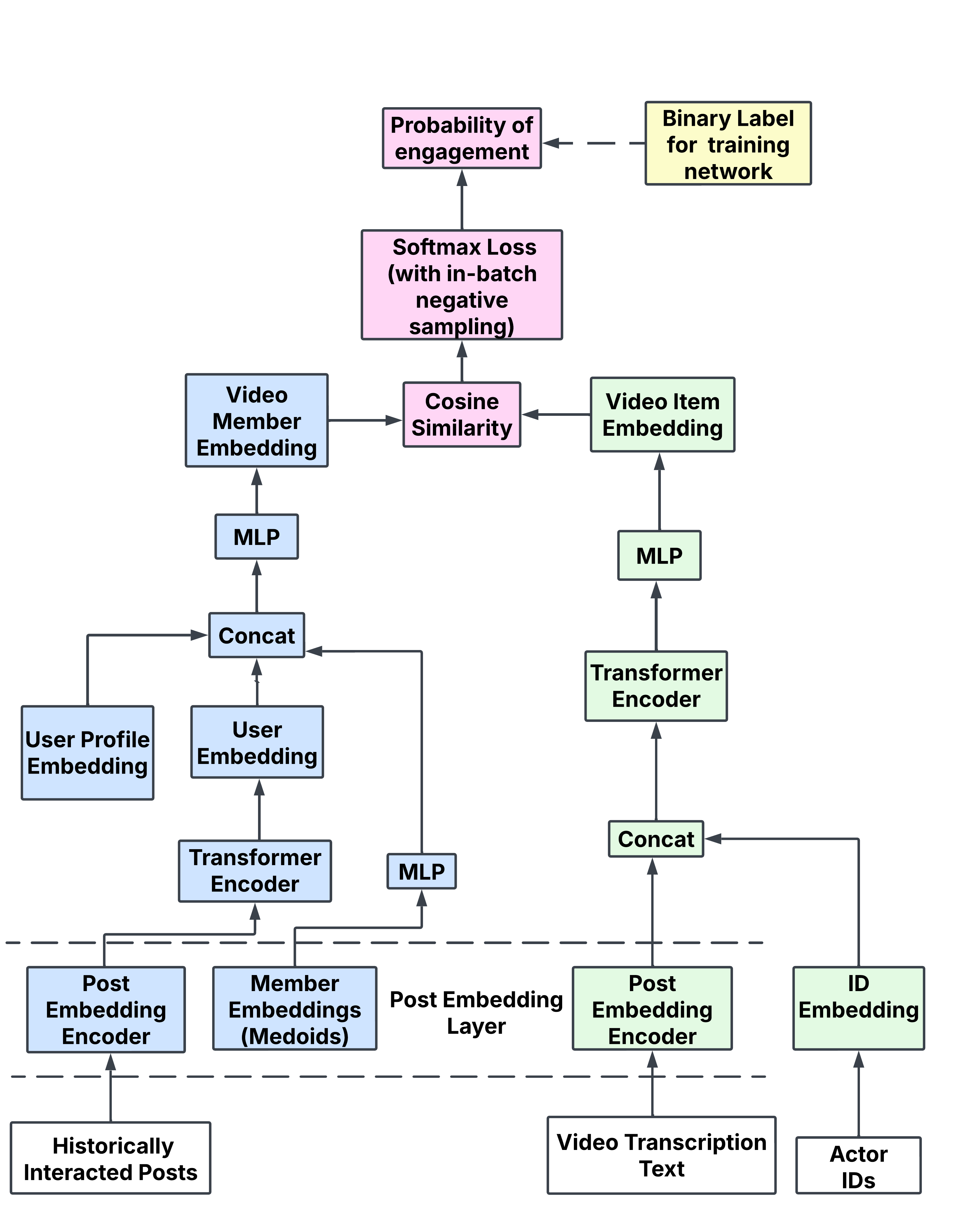}
    \caption{Video retrieval model architecture with LinkedIn post embeddings}
    \Description{Video retrieval model architecture with LinkedIn post embeddings}
    \label{fig:video_retrieval_model_arch}
\end{figure}

LinkedIn has a video recommendation experience across multiple product surfaces like Video Tab, Video Chaining and Video Carousel  where professional content in the format of short videos are available for our members. We added post embeddings based on video transcript information to the video retrieval layer (detailed architecture in \autoref{fig:video_retrieval_model_arch})  and we achieved an improvement of \textbf{+10.46\%} in \textbf{Total Watch Time} ($p < 0.0001$) and \textbf{+1.74\%} in \textbf{DAU} ($p < 0.0001$) for the video tab surface. The LinkedIn Post Embedding model was not retrained, and was simply used for inference using the video transcript text as input to generate embeddings. This further validates the ability of the embeddings to easily scale across multiple product surfaces with impact.

\section{Conclusion}

In this paper, we presented the fine-tuning of a BERT-based language model to generate high quality post embeddings, that have been widely adopted across LinkedIn. We demonstrated how these embeddings enable the derivation of member representations using hierarchical clustering (Ward’s method), and showed that training on a diverse set of semantic labeling tasks led to consistent performance improvements through positive transfer. The generalizability of the model was further validated through its zero-shot learning capabilities on an unseen task. We also compared the model’s performance against OpenAI’s generalized embeddings, highlighting its superior effectiveness for LinkedIn specific applications achieved at significantly lower dimensionality. Furthermore, we outlined several real world product use cases where these embeddings were deployed, resulting in measurable online impact through A/B testing. Looking ahead, our goal is to integrate larger foundational language models into our fine-tuning framework, and expand our training datasets by partnering with additional LinkedIn product teams. This ongoing effort towards platformization not only enhances the quality and scalability of post embeddings, but also incentivizes the creation of new, reusable datasets that can benefit multiple teams across the company. 

\section*{Limitations}
Our embeddings have been battle-tested in production for over two years and remain competitive, although advances in foundational and multimodal models may eventually surpass this approach. Our framework is designed to incorporate such improvements. We chose binary cross-entropy loss for efficiency: while triplet loss and InfoNCE are standard, triplet loss required three parallel inferences per update, increasing GPU memory costs and reducing batch sizes, which degraded performance. BCE provided a more practical trade-off between efficiency and downstream accuracy. Our evaluation metric (\textit{AvgFracTripletsWherePosIsCloser}) aligns with our semantic understanding objective, although alternative ranking or retrieval based offline metrics could provide complementary views of embedding quality. Finally, embedding performance is sensitive to how positive and negative pairs are sampled; although we adopted a broad, multi-task strategy, more sophisticated pair generation may further improve generalization. We have initiated efforts to develop multimodal post embeddings that integrate both visual and textual information, since the current embeddings only operate on the text present in a post.




\bibliographystyle{ACM-Reference-Format}
\balance
\bibliography{bibliography}


\begin{thebibliography}{12}


\ifx \showCODEN    \undefined \def \showCODEN     #1{\unskip}     \fi
\ifx \showISBNx    \undefined \def \showISBNx     #1{\unskip}     \fi
\ifx \showISBNxiii \undefined \def \showISBNxiii  #1{\unskip}     \fi
\ifx \showISSN     \undefined \def \showISSN      #1{\unskip}     \fi
\ifx \showLCCN     \undefined \def \showLCCN      #1{\unskip}     \fi
\ifx \shownote     \undefined \def \shownote      #1{#1}          \fi
\ifx \showarticletitle \undefined \def \showarticletitle #1{#1}   \fi
\ifx \showURL      \undefined \def \showURL       {\relax}        \fi
\providecommand\bibfield[2]{#2}
\providecommand\bibinfo[2]{#2}
\providecommand\natexlab[1]{#1}
\providecommand\showeprint[2][]{arXiv:#2}

\bibitem[Aghajanyan et~al\mbox{.}(2021)]%
        {aghajanyan2021muppet}
\bibfield{author}{\bibinfo{person}{Armen Aghajanyan}, \bibinfo{person}{Anchit Gupta}, \bibinfo{person}{Akshat Shrivastava}, \bibinfo{person}{Xilun Chen}, \bibinfo{person}{Luke Zettlemoyer}, {and} \bibinfo{person}{Sonal Gupta}.} \bibinfo{year}{2021}\natexlab{}.
\newblock \showarticletitle{Muppet: Massive multi-task representations with pre-finetuning}.
\newblock \bibinfo{journal}{\emph{arXiv preprint arXiv:2101.11038}} (\bibinfo{year}{2021}).
\newblock


\bibitem[Borisyuk et~al\mbox{.}(2024a)]%
        {borisyuk2024lignn}
\bibfield{author}{\bibinfo{person}{Fedor Borisyuk}, \bibinfo{person}{Shihai He}, \bibinfo{person}{Yunbo Ouyang}, \bibinfo{person}{Morteza Ramezani}, \bibinfo{person}{Peng Du}, \bibinfo{person}{Xiaochen Hou}, \bibinfo{person}{Chengming Jiang}, \bibinfo{person}{Nitin Pasumarthy}, \bibinfo{person}{Priya Bannur}, \bibinfo{person}{Birjodh Tiwana}, {et~al\mbox{.}}} \bibinfo{year}{2024}\natexlab{a}.
\newblock \showarticletitle{LiGNN: Graph Neural Networks at LinkedIn}. In \bibinfo{booktitle}{\emph{Proceedings of the 30th ACM SIGKDD Conference on Knowledge Discovery and Data Mining}}. \bibinfo{pages}{4793--4803}.
\newblock


\bibitem[Borisyuk et~al\mbox{.}(2024b)]%
        {borisyuk2024lirank}
\bibfield{author}{\bibinfo{person}{Fedor Borisyuk}, \bibinfo{person}{Mingzhou Zhou}, \bibinfo{person}{Qingquan Song}, \bibinfo{person}{Siyu Zhu}, \bibinfo{person}{Birjodh Tiwana}, \bibinfo{person}{Ganesh Parameswaran}, \bibinfo{person}{Siddharth Dangi}, \bibinfo{person}{Lars Hertel}, \bibinfo{person}{Qiang~Charles Xiao}, \bibinfo{person}{Xiaochen Hou}, {et~al\mbox{.}}} \bibinfo{year}{2024}\natexlab{b}.
\newblock \showarticletitle{LiRank: Industrial Large Scale Ranking Models at LinkedIn}. In \bibinfo{booktitle}{\emph{Proceedings of the 30th ACM SIGKDD Conference on Knowledge Discovery and Data Mining}}. \bibinfo{pages}{4804--4815}.
\newblock


\bibitem[Brown et~al\mbox{.}(2020)]%
        {brown2020language}
\bibfield{author}{\bibinfo{person}{Tom Brown}, \bibinfo{person}{Benjamin Mann}, \bibinfo{person}{Nick Ryder}, \bibinfo{person}{Melanie Subbiah}, \bibinfo{person}{Jared~D Kaplan}, \bibinfo{person}{Prafulla Dhariwal}, \bibinfo{person}{Arvind Neelakantan}, \bibinfo{person}{Pranav Shyam}, \bibinfo{person}{Girish Sastry}, \bibinfo{person}{Amanda Askell}, {et~al\mbox{.}}} \bibinfo{year}{2020}\natexlab{}.
\newblock \showarticletitle{Language models are few-shot learners}.
\newblock \bibinfo{journal}{\emph{Advances in neural information processing systems}}  \bibinfo{volume}{33} (\bibinfo{year}{2020}), \bibinfo{pages}{1877--1901}.
\newblock


\bibitem[Conneau et~al\mbox{.}(2019)]%
        {conneau2019unsupervised}
\bibfield{author}{\bibinfo{person}{Alexis Conneau}, \bibinfo{person}{Kartikay Khandelwal}, \bibinfo{person}{Naman Goyal}, \bibinfo{person}{Vishrav Chaudhary}, \bibinfo{person}{Guillaume Wenzek}, \bibinfo{person}{Francisco Guzm{\'a}n}, \bibinfo{person}{Edouard Grave}, \bibinfo{person}{Myle Ott}, \bibinfo{person}{Luke Zettlemoyer}, {and} \bibinfo{person}{Veselin Stoyanov}.} \bibinfo{year}{2019}\natexlab{}.
\newblock \showarticletitle{Unsupervised cross-lingual representation learning at scale}.
\newblock \bibinfo{journal}{\emph{arXiv preprint arXiv:1911.02116}} (\bibinfo{year}{2019}).
\newblock


\bibitem[Devlin(2018)]%
        {devlin2018bert}
\bibfield{author}{\bibinfo{person}{Jacob Devlin}.} \bibinfo{year}{2018}\natexlab{}.
\newblock \showarticletitle{Bert: Pre-training of deep bidirectional transformers for language understanding}.
\newblock \bibinfo{journal}{\emph{arXiv preprint arXiv:1810.04805}} (\bibinfo{year}{2018}).
\newblock


\bibitem[Greene et~al\mbox{.}(2022)]%
        {greene2022new}
\bibfield{author}{\bibinfo{person}{Ryan Greene}, \bibinfo{person}{Ted Sanders}, \bibinfo{person}{Lilian Weng}, {and} \bibinfo{person}{Arvind Neelakantan}.} \bibinfo{year}{2022}\natexlab{}.
\newblock \showarticletitle{New and improved embedding model}.
\newblock \bibinfo{journal}{\emph{OpenAI Blog. Available online: https://openai. com/blog/new-and-improved-embedding-model (accessed on 28 November 2023)}} (\bibinfo{year}{2022}).
\newblock


\bibitem[Liu(2019)]%
        {liu2019roberta}
\bibfield{author}{\bibinfo{person}{Yinhan Liu}.} \bibinfo{year}{2019}\natexlab{}.
\newblock \showarticletitle{Roberta: A robustly optimized bert pretraining approach}.
\newblock \bibinfo{journal}{\emph{arXiv preprint arXiv:1907.11692}}  \bibinfo{volume}{364} (\bibinfo{year}{2019}).
\newblock


\bibitem[Pal et~al\mbox{.}(2020)]%
        {pal2020pinnersage}
\bibfield{author}{\bibinfo{person}{Aditya Pal}, \bibinfo{person}{Chantat Eksombatchai}, \bibinfo{person}{Yitong Zhou}, \bibinfo{person}{Bo Zhao}, \bibinfo{person}{Charles Rosenberg}, {and} \bibinfo{person}{Jure Leskovec}.} \bibinfo{year}{2020}\natexlab{}.
\newblock \showarticletitle{Pinnersage: Multi-modal user embedding framework for recommendations at pinterest}. In \bibinfo{booktitle}{\emph{Proceedings of the 26th ACM SIGKDD International Conference on Knowledge Discovery \& Data Mining}}. \bibinfo{pages}{2311--2320}.
\newblock


\bibitem[Reimers(2019)]%
        {reimers2019sentence}
\bibfield{author}{\bibinfo{person}{N Reimers}.} \bibinfo{year}{2019}\natexlab{}.
\newblock \showarticletitle{Sentence-BERT: Sentence Embeddings using Siamese BERT-Networks}.
\newblock \bibinfo{journal}{\emph{arXiv preprint arXiv:1908.10084}} (\bibinfo{year}{2019}).
\newblock


\bibitem[Wang et~al\mbox{.}(2022)]%
        {wang2022text}
\bibfield{author}{\bibinfo{person}{Liang Wang}, \bibinfo{person}{Nan Yang}, \bibinfo{person}{Xiaolong Huang}, \bibinfo{person}{Binxing Jiao}, \bibinfo{person}{Linjun Yang}, \bibinfo{person}{Daxin Jiang}, \bibinfo{person}{Rangan Majumder}, {and} \bibinfo{person}{Furu Wei}.} \bibinfo{year}{2022}\natexlab{}.
\newblock \showarticletitle{Text embeddings by weakly-supervised contrastive pre-training}.
\newblock \bibinfo{journal}{\emph{arXiv preprint arXiv:2212.03533}} (\bibinfo{year}{2022}).
\newblock


\bibitem[Zoupanos et~al\mbox{.}(2022)]%
        {zoupanos2022efficient}
\bibfield{author}{\bibinfo{person}{Spyros Zoupanos}, \bibinfo{person}{Stratis Kolovos}, \bibinfo{person}{Athanasios Kanavos}, \bibinfo{person}{Orestis Papadimitriou}, {and} \bibinfo{person}{Manolis Maragoudakis}.} \bibinfo{year}{2022}\natexlab{}.
\newblock \showarticletitle{Efficient comparison of sentence embeddings}. In \bibinfo{booktitle}{\emph{Proceedings of the 12th Hellenic Conference on Artificial Intelligence}}. \bibinfo{pages}{1--6}.
\newblock


\end{thebibliography}

\end{document}